\documentclass[letterpaper, 10 pt, conference]{ieeeconf}  %

\IEEEoverridecommandlockouts                              %

\usepackage{amsmath,amssymb,amsfonts}
\usepackage{algorithmic}
\usepackage{graphicx}
\usepackage{textcomp}
\usepackage{xcolor}
\usepackage{siunitx}
\usepackage{hyperref}
\usepackage{placeins}
\usepackage{flushend}
\usepackage[caption=false,font=small]{subfig}
\usepackage{cite}
\usepackage{tikz}

\def\BibTeX{{\rm B\kern-.05em{\sc i\kern-.025em b}\kern-.08em
    T\kern-.1667em\lower.7ex\hbox{E}\kern-.125emX}}

\usepackage{etoolbox}
\makeatletter
\patchcmd{\@makecaption}
  {\scshape}
  {}
  {}
  {}
\makeatletter
\patchcmd{\@makecaption}
  {\\}
  {.\ }
  {}
  {}
\makeatother
\def\tablename{Table}

\title{\LARGE \bf
DeepSTEP - Deep Learning-Based Spatio-Temporal End-To-End Perception for Autonomous Vehicles
}

\author{Sebastian Huch$^{1,*}$, Florian Sauerbeck$^{1,*}$, and Johannes Betz$^{2}$ %
\thanks{$^{*}$Authors contributed equally.
        }%
\thanks{$^{1}$Sebastian Huch and Florian Sauerbeck are with the Institute of Automotive Technology, School of Engineering \& Design,
        Technical University of Munich, 85748 Garching, Germany \newline
        {\tt\small \{sebastian.huch, florian.sauerbeck\}@tum.de}}%
\thanks{$^{2}$Johannes Betz is with the Professorship Autonomous Vehicle Systems, Technical University of Munich, 85748 Garching, Germany
        {\tt\small johannes.betz@tum.de}}%
\thanks{The research was partially funded by the Federal Ministry of Education and Research of Germany (BMBF) within the project Wies'n Shuttle (FKZ 03ZU1105AA) in the MCube cluster, and through basic research funds from the Institute for Automotive Technology.}
}

\newcommand\copyrighttext{%
    \footnotesize \textcopyright 2023 IEEE.  Personal use of this material is permitted.  Permission from IEEE must be obtained for all other uses, in any current or future media, including reprinting/republishing this material for advertising or promotional purposes, creating new collective works, for resale or redistribution to servers or lists, or reuse of any copyrighted component of this work in other works.
}
\newcommand\copyrightnotice{%
    \begin{tikzpicture}[remember picture,overlay]
    \node[anchor=south,yshift=0pt, xshift=10pt] at (current page.south) {\fbox{\parbox{\dimexpr\textwidth-\fboxsep-\fboxrule\relax}{\copyrighttext}}};
    \end{tikzpicture}%
}

\begin{document}

\maketitle
\copyrightnotice

\begin{abstract}
Autonomous vehicles demand high accuracy and robustness of perception algorithms.
To develop efficient and scalable perception algorithms, the maximum information should be extracted from the available sensor data.
In this work, we present our concept for an end-to-end perception architecture, named \textit{DeepSTEP}.
The deep learning-based architecture processes raw sensor data from the camera, LiDAR, and RaDAR, and combines the extracted data in a deep fusion network.
The output of this deep fusion network is a shared feature space, which is used by perception head networks to fulfill several perception tasks, such as object detection or local mapping.
\textit{DeepSTEP} incorporates multiple ideas to advance state of the art:
First, combining detection and localization into a single pipeline allows for efficient processing to reduce computational overhead and further improves overall performance.
Second, the architecture leverages the temporal domain by using a self-attention mechanism that focuses on the most important features.
We believe that our concept of \textit{DeepSTEP} will advance the development of end-to-end perception systems.
The network will be deployed on our research vehicle, which will be used as a platform for data collection, real-world testing, and validation.
In conclusion, \textit{DeepSTEP} represents a significant advancement in the field of perception for autonomous vehicles.
The architecture's end-to-end design, time-aware attention mechanism, and integration of multiple perception tasks make it a promising solution for real-world deployment.
This research is a work in progress and presents the first concept of establishing a novel perception pipeline.
\end{abstract}

\section{Introduction}
\begin{figure}[t!]
	\centering
            \subfloat[Modular Pipeline\label{fig:modular_pipeline}]
            {\includegraphics[width=1.0\linewidth]{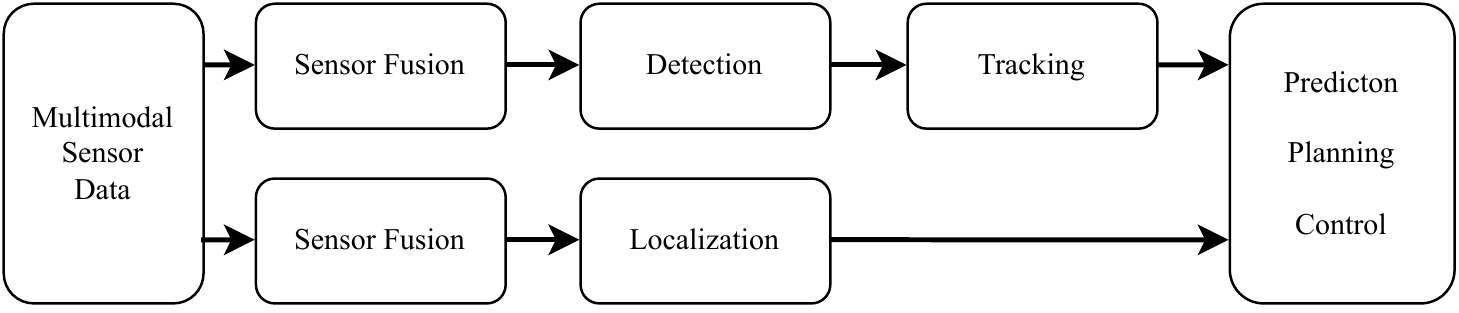}}
            \hfill
            \subfloat[DeepSTEP\label{fig:high_level_overview}]
            {\includegraphics[width=1.0\linewidth]{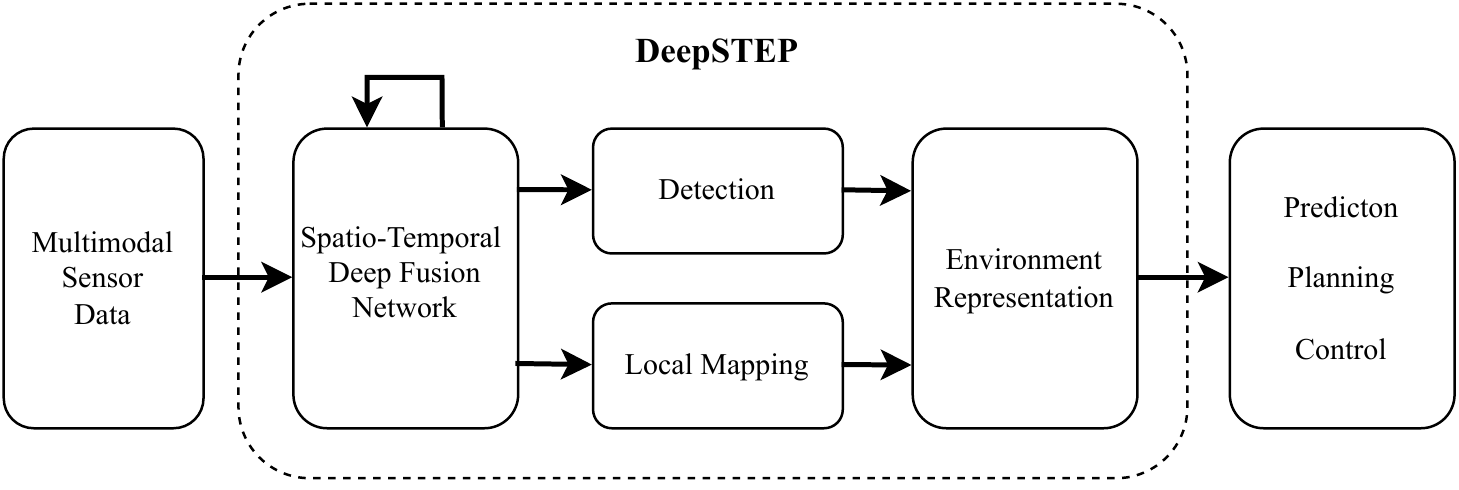}}
		\caption{Comparison of the modular pipeline and \textit{DeepSTEP} to solve the perception task for autonomous vehicles.}
		\label{fig:overview}
\end{figure}

Software plays a key role in autonomous driving.
A common architecture to approach the task of autonomous driving is a modular pipeline consisting of sequential modules of perception, prediction, planning, and control \cite{pendlton2017}, as depicted in Fig. \ref{fig:modular_pipeline}.
Although research is conducted in all areas of the pipeline, specifically perception algorithms are a crucial part of the software pipeline and limit today’s vehicles to reaching higher levels of autonomy \cite{fayyad2020sensorfusion}.
Perception covers multiple tasks including, but not limited to, object detection of traffic participants and infrastructure, and localization and mapping of the ego vehicle's environment.
The algorithms use sensor data from the camera, LiDAR, and RaDAR sensors to solve these tasks.
To advance technology and increase the level of autonomy, more robust and reliable perception algorithms must be developed.

The input for the perception algorithms is usually only a single frame, i.e. one image or point cloud, or more frames for architectures using recurrent neural networks or transformer networks.
In the case of using a single frame as input, the networks are not aware of time and drop previously extracted information.
State-of-the-art autonomous driving software stacks cope with this challenge by adding an additional probabilistic tracking algorithm.
These tracking algorithms track and estimate the position of an object even if it is completely occluded for several seconds and, therefore, can include the time dimension.
They are based on physical assumptions and require extensive manual parameter tuning, which is time-consuming, hardly scalable, error-prone, and always leads to a trade-off between performance and reliability.

Furthermore, in current state-of-the-art perception stacks of autonomous vehicles, object detection, mapping, and localization tasks are handled as individual modules.
Information is not shared between these tasks, even though both tasks use the same input data and extract similar information from the input data.
Therefore, these tasks are spatially decoupled and cannot benefit from the local features extracted by the other algorithm, leading to computational overhead and information loss.

To address the above-mentioned \textbf{research gap} and strive towards a scalable, real-world applicable perception stack, we present a deep learning-based spatio-temporal end-to-end perception concept called \textit{DeepSTEP}.
The overall concept is depicted in Fig. \ref{fig:high_level_overview}.

The \textbf{novel contributions} of \textit{DeepSTEP} can be summarized as follows:
\begin{itemize}
    \item We propose an end-to-end perception stack with a shared feature space for multiple perception tasks, named \textit{DeepSTEP}.
    \item \textit{DeepSTEP} incorporates the temporal domain in the feature space to maximize the information content.
    \item \textit{DeepSTEP} improves the efficiency and runtime using a shared feature extractor for detection and local mapping.
    \item \textit{DeepSTEP} is expandable by integrating new perception heads.
    \item We propose an efficient, holistic representation of the entire perceived environment for the succeeding modules in the AV software stack.
\end{itemize}

\section{Related Work}
\label{sec:related_work}
Research in the area of spatio-temporal end-to-end perception is limited.
The majority of research focuses on improving the state-of-the-art algorithms found in the decomposed modular approach, such as new or optimized object detection algorithms, more efficient localization and local mapping techniques, or improved tracking algorithms.
In the following, we will first review the state-of-the-art algorithms of the modular approach, which form a crucial baseline for \textit{DeepSTEP}, before discussing end-to-end perception approaches.
Since our proposed network \textit{DeepSTEP} further incorporates sensor fusion with feature extraction and an efficient environment representation, we will also cover the state of the art in these research areas.

\vspace{0.5em}
\textbf{Sensor fusion} is the process of merging data from multiple sources, which can be homogeneous or heterogeneous sensor modalities such as camera, LiDAR, and RaDAR \cite{fayyad2020sensorfusion}.
The goal of sensor fusion is to increase the robustness and performance of perception, as well as to overcome the limitations of individual sensor modalities.
Sensor fusion can be categorized into three approaches: early fusion, mid-level fusion, and late fusion \cite{kolar2020sensorfusion}.

Early fusion operates on raw sensor data, e.g. with a fusion of camera images and LiDAR point clouds.
It is usually focused on camera and LiDAR fusion using deep learning algorithms to process the fused data \cite{qi2017frustumpointnet, liang2020sensorfusion}.
The first approaches to camera and RaDAR with various levels of fusion from early to mid-level fusion have been developed by \cite{nobis2020crf}.

Mid-level fusion combines data based on the feature level.
Individual algorithms extract abstract features for the different sensor modalities, which are later combined and processed jointly to obtain the final perception output.
In \cite{almalioglu2022deep}, the authors present a deep learning-based sensor fusion approach for a localization application. Ego motion and pixel-wise depth are estimated with multi-modal sensor data. A fusion net fuses the unaligned motion predictions. An additional geometric transformer is used to reconstruct the 3D scene.

Late fusion is an approach that fuses the data at the decision level.
Specifically, the sensor data are processed by individual algorithms for each sensor modality, and the final outputs of each algorithm are fused in the following step.
Late fusion has the advantage of being robust to hardware or software errors of a single modality.
However, the algorithms are not specialized to directly learn the strengths and weaknesses of the individual modalities.

\vspace{0.5em}
\textbf{Object Detection.} In recent years, object detection research has made substantial progress thanks to deep neural networks and the availability of large public data sets, such as KITTI \cite{geiger2013vision}, Waymo Open Dataset \cite{Sun_2020_CVPR}, or nuScenes \cite{nuscenes}.
These data sets include challenges to benchmark the object detection algorithms developed, further motivating research in this area.
Large-scale object detection started a decade ago with convolutional neural networks (CNNs) that extract 2D object positions on camera images.
Following 2D object detection, also research on 3D object detection using camera images as well as LiDAR point clouds progressed fast.

Camera-based 3D object detection is a very active research topic, with multiple works published every year \cite{brazil2019m3drpn}, \cite{wang2018pseudolidar}.
One of the main challenges is the recovery of the object's 3D location after the 3D space has been projected onto a 2D image.
Approaches can be categorized into monocular or stereo-based algorithms, which use a single camera or a pair of cameras, respectively.
To recover the 3D information of the objects in the image, monocular object detection algorithms make use of predefined anchors, depth images, or object priors.
The M3D-RPN network \cite{brazil2019m3drpn} regresses the 3D bounding box parameters for each predefined anchor placed on the image.
Pseudo-LiDAR \cite{wang2018pseudolidar} is a network architecture that first calculates a depth image with depth values for each image pixel.
This depth image can be represented as a "pseudo" 3D point cloud, which further serves as input for conventional LiDAR 3D object detectors.
Other approaches for 3D camera object detection are based on object priors, such as distinctive object shapes, that are regressed during inference.
For example, Mono3D++ \cite{he2019mono3dpp} uses morphable wireframe models as shape priors to detect vehicles in camera images.

Research on LiDAR-based 3d object detection algorithms started in 2015, and since then several works have been published each year, pushing the performance with every new approach.
Notable approaches include point-based PointRCNN \cite{shi2018pointrcnn}, pillar-based PointPillars \cite{lang2018pointpillars}, point-voxel-based PV-RCNN \cite{shi2019pvrcnn}, or recent transformer-based algorithms, such as Voxel Transformer \cite{mao2021voxeltransformer}.
Incorporating recurrent neural networks or transformer networks enables the use of multiple input frames, which allows 3D object detection networks to be time-aware.
Notable approaches include \cite{huang2020lidarlstm} using LSTM, \cite{yuan2022tctr} using transformers, or \cite{yang20213dman} using temporal attention.

When comparing the performance of camera- and LiDAR-based 3D object detection, especially monocular camera methods perform much worse than LiDAR methods, since 3D information is lost on single 3D images. 
For an extensive study of 3D object detection, we refer the reader to \cite{mao2022surveyobjectdetection}.

\vspace{0.5em}
\textbf{Localization and Local Mapping.} Currently, most autonomous driving approaches rely on precise high-definition maps (HD maps).
The biggest challenge with this approach is poor scalability for several reasons \cite{li2022hdmapnet}.
To address these problems, research has been conducted to develop more scalable HD mapping approaches.
Qin et al. \cite{qin2021light} propose a method for creating lane-level maps with dedicated mapping vehicles.
HDMapNet \cite{li2022hdmapnet} was implemented to generate HD maps by fusing LiDAR and camera observations.
Manual refinement of the generated maps was necessary.
However, a precise location within a global map is not necessary for normal traffic driving tasks \cite{meyer2018deep}.
A precise representation of the local environment and the location of the ego within is more important and can open the door for more scalable approaches.
Therefore, research has arisen in the field of novel environment perception.
The most recent approaches use 2D semantic maps in birds-eye-view (BEV) as a map format.
SMNet \cite{cartillier2021semantic} uses an RGB-D camera and assumes known localization to create semantic maps of unknown environments.
Due to the limited range and high light dependency of the RGB-D camera, the application of the approach is limited to indoor environments.
In contrast, GitNet \cite{gong2022gitnet} was particularly developed for an automotive use case.
Similar approaches have been introduced in \cite{xie2022m} and \cite{roddick2020predicting}.
Multiple cameras are fused together in these for maximum field of view.
With the use of Transformer networks, a BEV image of the vehicle surroundings is generated and subsequently semantically segmented to receive a semantic map in BEV.
Some of the approaches included Bayesian filters to fuse single frames using temporal relations.
All authors state that this temporal fusion can show a significant improvement in the results and should be investigated in future works.
Saha et al. \cite{saha2021enabling} proposed a dedicated method for the spatio-temporal aggregation of BEV information.
It could be shown that a fusion in the BEV plane outperforms a fusion in the image plane.
BEVFormer \cite{li2022bevformer} uses spatio-temporal transformers to predict BEV semantic maps.
Therefore, spatial cross-attention and temporal self-attention are used.
A summary of the current state-of-the-art of BEV perception is given in \cite{li2022delving}.

One can see that the integration of the temporal domain into local semantic mapping is an ongoing research field.
Moreover, the integration of multi-modal sensor data will have to be addressed in future work.

\vspace{0.5em}
\textbf{Environment Representation.} The field of environment representation is a major research field.
The above-mentioned approaches use 2D BEV semantic maps to represent the local environment of the ego vehicle.
Some of them also include dynamic objects; some only depict the static environment.

Recently, progress in the field of neural networks has enabled new possibilities for precise reconstruction and representation of the environment.
Particularly Neural Radiance Fields (NeRFs) \cite{mildenhall2021nerf} are to be mentioned.
Deng et al. \cite{deng2022depth} present a methodology for combined perception tasks using NeRFs.
No training data and fewer images are needed compared to other approaches.
CLONeR (Camera-Lidar Fusion for Occupancy Grid-Aided Neural Representation) \cite{carlson2022cloner} is an application of neural representations for outdoor autonomous driving scenes.
A 3D occupancy grid map is used to represent the environment.
Block-NeRF \cite{tancik2022block} and BungeeNeRF \cite{xiangli2022bungeenerf} tackle the challenge of scalable NeRFs for city-scale representations.
Loc-NeRF \cite{maggio2022loc} and NeRF-SLAM \cite{rosinol2022nerf} demonstrated the feasibility of using NeRFs for the localization and mapping of autonomous robots.

Dense environment reconstructions can be used to extract maximum information and generate an efficient environment representation for the path planning of the autonomous vehicle.
Neural scene graphs \cite{rosinol20203d, ost2021neural} present a promising way to realize this.
Neural networks are used to implicitly represent dynamic multi-object scenes.
The information contained in the graph built from perception sensor data can directly be used by path planning, e.g. as it contains all relevant information.
The dependency on the database and thus the scalability of these approaches will have to be researched.

\begin{figure*}[ht!]
    \centering
        \includegraphics[width=0.95\linewidth, trim={0cm 0cm 0cm 0cm}, clip]{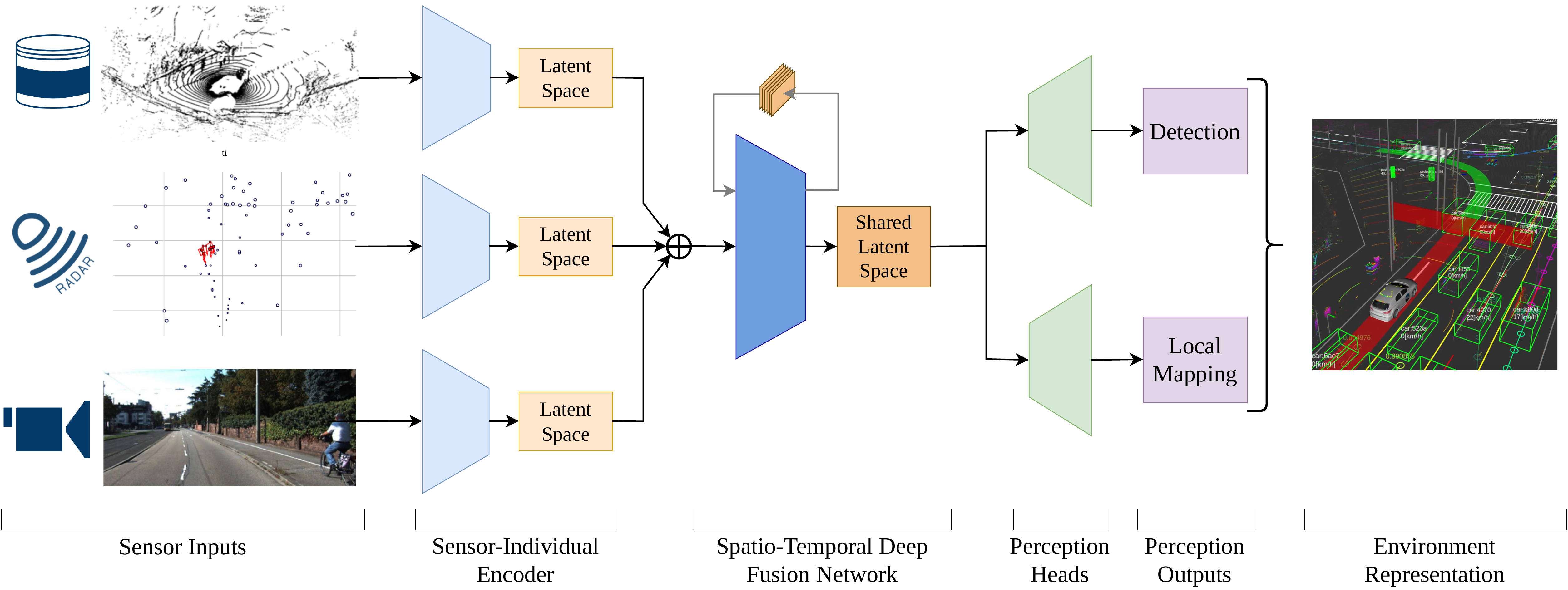}
        \caption{Network structure of \textit{DeepSTEP} from sensor inputs to perception outputs. These will be integrated into the subsequent environment representation.}
        \label{fig:network}
\end{figure*}

\vspace{0.5em}
\textbf{End-to-End Perception.} The progress in the area of detection and localization is tremendous, with several hundred works published each year.
However, the combination of these algorithms to form an end-to-end perception covering spatial and temporal domains is rarely studied.
Unlike complete end-to-end autonomous driving \cite{huch2021end2end}, which covers perception, prediction, planning, and control in an end-to-end manner, end-to-end perception has the benefit of fixed system boundaries and interpretable network output.

\cite{liang2020pnpnet} propose PnPNet, which is an end-to-end pipeline for joint perception and prediction.
The goal of this pipeline is to detect and track objects, as well as predict the future motion of these objects surrounding the ego vehicle.
The authors state that PnPNet achieves better occlusion recovery and prediction results than comparable state-of-the-art approaches.
Multiple aggregated LiDAR sweeps serve as input to the network.
Although this architecture covers multiple perception tasks and combines the spatial and temporal domains, it focuses solely on surrounding objects and does not incorporate the ego vehicle's localization.
HybridNets \cite{vu2022hybridnets} is a multi-task network designed for object detection, drivable space segmentation, and lane detection, and therefore covers detection and localization tasks.
The network consists of one encoder and two network heads, specifically one detection head and one segmentation head.
The inputs consist of single images from a front-view camera, and therefore HybridNets neglects the temporal domain.
Another approach to end-to-end perception is ST-P3 \cite{hu2022stp3}.
This visual-based network uses spatial and temporal domains to learn features for perception, prediction, and path planning.
The authors report their results on an open-loop dataset as well as on a closed-loop simulation.
ST-3P shows superior performance on the selected benchmarks.

All of the described works of end-to-end perception pick up the idea of combining either spatial and temporal domains or combining the tasks of detection and localization.
However, none of these works exploits the combination of a spatio-temporal domain and multi-task learning for joint detection and localization.
Furthermore, these approaches make use of only a single sensor modality, either camera or LiDAR.

\section{DeepSTEP}

Based on the open research topics presented above and the related work conducted, in the following, we present our end-to-end perception \textit{DeepSTEP}.
First, we will briefly describe the concept at a high level.
Second, we will present the four key elements of \textit{DeepSTEP}, namely deep fusion, detection, local mapping, and environment representation, in detail and highlight our contributions in each area.
Finally, we introduce the vehicle platform, as well as the data concept used for our proposed approach.

\subsection{Overview}

The goal of \textit{DeepSTEP} is to develop and extend current perception algorithms.
These deep learning-based algorithms should efficiently extract information from multi-modal sensor data, such as LiDAR, camera, and RaDAR, while being spatial- and time-aware, without the need for an additional tracking algorithm.
The information extracted from the available sensor data should be maximized by keeping the information through a certain time interval so that the algorithms benefit from the information extracted in the previous time steps.
Due to the shared feature space and joint training of all tasks, each network head can access all available information.
For example, this allows the network to learn that the probability of the existence of vehicles is higher on the roads than directly next to buildings.
Inversely, the probability of being next to a red light is higher if many static vehicles are detected.
Thus, all perception tasks can benefit from each other and computational resources can be saved.

Fig. \ref{fig:network} gives a detailed overview of the general concept of \textit{DeepSTEP}.
Our proposed approach combines research in the following four key areas:
\begin{itemize}
    \item Sensor-individual feature extraction and spatio-temporal deep sensor fusion;
    \item Detection using abstract features;
    \item Local mapping of the surrounding environment for mapless driving;
    \item Efficient environment representation.
\end{itemize}

In detail, the sensor data collected by the camera, LiDAR, and RaDAR are first processed by individual sensor encoders.
The latent features generated by these sensor individual encoders are concatenated.
The concatenated latent spaces are jointly processed in a deep fusion network to generate a shared latent space.
The deep fusion network includes a self-attention mechanism to process temporal information.
Therefore, the abstract shared latent space not only contains the spatial information of the AV's surrounding environment but also includes temporal information from previous timesteps.
Using this abstract shared feature space, the following task-specific perception heads predict the output for object detection and local mapping, such as objects and road-level information.

\subsection{Detailed Network Structure}
After briefly introducing the overall structure of \textit{DeepSTEP}, we present the individual key network blocks and our specific contributions in each block in the following.

\vspace{0.5em}
\textbf{Sensor Fusion}.
We aim to achieve feature-level sensor fusion.
This allows the abstraction of all sensor modalities to the same level where they are fused.
Compared to early-fusion or late-fusion approaches, no information is lost, and yet sensor fusion can be done in an efficient way.
Features will be extracted with sensor-individual encoders, and sensor fusion is handled by a deep fusion network that outputs a shared feature space.
The sensor-individual encoders will be based on network layers used by networks of the modular approach, e.g. convolutional layers for images or multi-layer perceptrons (MLPs) for point clouds.
The temporal domain will be integrated into the deep fusion network, which uses a self-attention mechanism known from transformer networks.
The shared feature space is used for subsequent perception tasks, such as object detection and local mapping.
Due to this modular setup, additional perception tasks based on this feature space can be integrated.

\vspace{0.5em}
\textbf{Object Detection}.
The goal of \textit{DeepSTEP}'s object detection module is to develop an algorithm capable of extracting objects from an abstract feature representation.
State-of-the-art approaches for object detection in the domain of autonomous driving use raw sensor data (images or point clouds) as input for deep neural networks as described in Sec. \ref{sec:related_work}.
The main difference between \textit{DeepSTEP}'s and modular approaches' object detection is that, instead of raw sensor data, it processes the shared feature space of the previous spatio-temporal deep fusion network.
Furthermore, it incorporates the temporal dimension, meaning that the input features include information from several multi-modal frames.
These characteristics lead to new requirements for the object detection algorithm, which should be based on existing work.
The output of the object detection algorithm is a list of objects that contains the position, orientation, size, and type of each detected object.
This output is combined with the output of the local mapping to form the global environment representation.

To summarize the contributions in the field of object detection, the following additions to the current state of the art will be made:
\begin{itemize}
    \item Use of input based on previously extracted high-level features from a spatio-temporal deep fusion network;
    \item Input contains information from several multi-model frames.
\end{itemize}

\begin{figure*}[ht!]
	\centering
		\includegraphics[width=0.95\linewidth, trim={0cm 0cm 0cm 0cm}, clip]{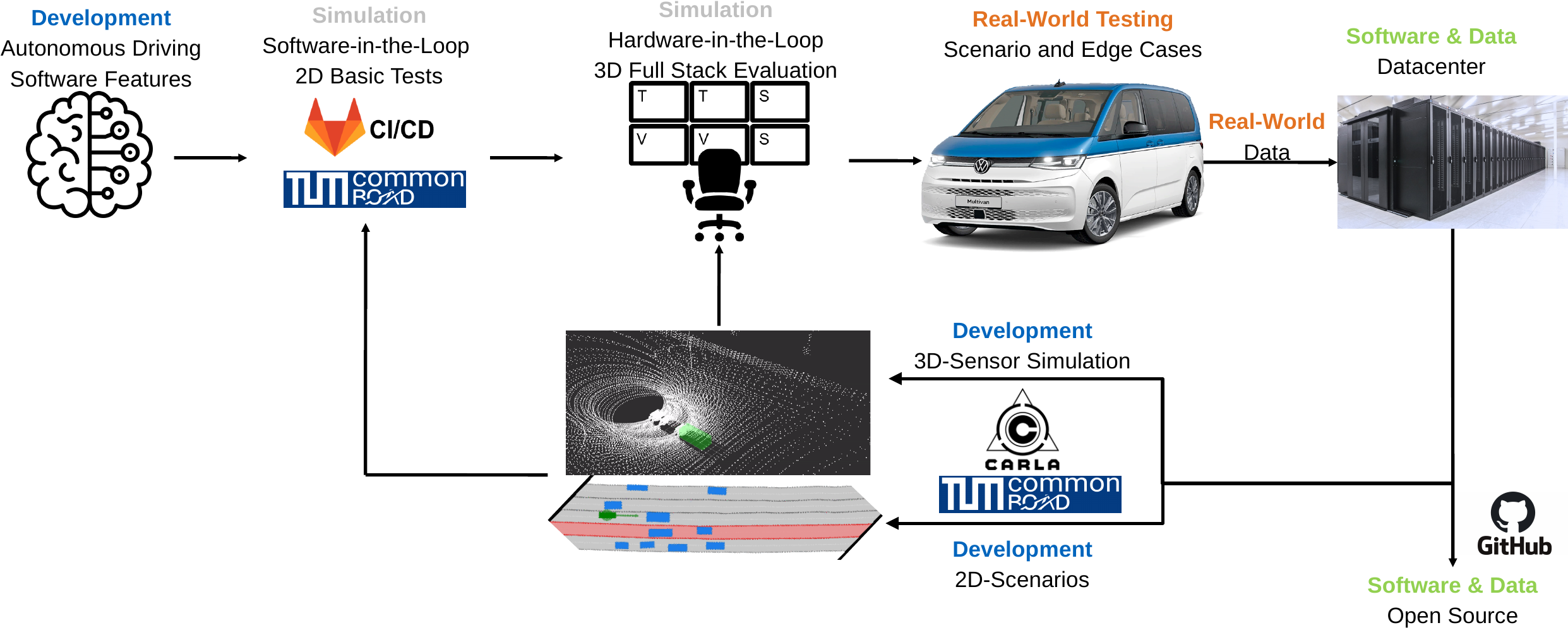}
		\caption{Overview of the development and testing workflow that will be applied within DeepSTEP.}
		\label{fig:dev_workflow}
\end{figure*}

\vspace{0.5em}
\textbf{Local Mapping}.
To enable level 5 autonomous driving, scalable approaches are needed for map handling \cite{betz2019whatcanwelearn}.
Therefore, we want to focus on the scalability and generalization of the algorithms developed.
In the long term, this could make the use of HD maps unnecessary to enable level 5 autonomous driving.

Our approach combines a local mapping module for the static environment with a dedicated object detection module for dynamic objects.
Therefore, a shared feature space is used.
In comparison to existing approaches, we do not want to fuse the results of single mapping frames but already include the temporal domain into the feature space and thus use this information in each frame's mapping step.
Moreover, we also plan to include different sensor modalities, namely camera, LiDAR, and RaDAR, to allow for maximum extraction of environmental information.
In addition, sensor failures and harsh conditions, such as bad weather, can be handled more robustly, and thus more safely.
The local mapping will output all relevant information for the subsequent modules, such as scene understanding and path planning.
Therefore, road-level information will be extracted from sensor data.
Not only will 2D BEV mapping be evaluated, but also 3D approaches that try to reconstruct the surrounding environment based on sensor data.

To summarize the contributions in the field of local mapping, the following additions to the current state-of-the-art will be made:
\begin{itemize}
    \item Local approach without the use of HD maps;
    \item Use of abstract features from different sensor modalities;
    \item Integration of the temporal domain into the local mapping step without a separate probabilistic fusion algorithm.
\end{itemize}

\vspace{0.5em}
\textbf{Environment Representation}.
An efficient environment representation is needed to provide the generated perception information to subsequent software modules, such as scene understanding or path planning.
This representation should include all relevant information, including the static environment, dynamic objects, and possibly trajectory predictions of dynamic objects.
The environment representation acts as an interface between the \textit{DeepSTEP} perception module and the subsequent autonomy stack; compare Fig. \ref{fig:high_level_overview}.

Local semantic mapping approaches using BEV use a 2D frame to represent the environment of the ego vehicle.
Within the scope of \textit{DeepSTEP}, 3D representations will also be evaluated.
Currently, the most promising approach for this is NeRFs.
Using multiple cameras and laser scanners of the autonomous vehicle, a precise 3D reconstruction of the environment can be generated.
As demonstrated in \cite{ZhiICCV2021}, these 3D representations can also include semantic labels, which are essential for path planning.
To incorporate everything in an efficient representation, neural scene graphs could be an exciting concept for handling all perception outputs.

Novel contributions to environmental representations of autonomous vehicles are:
\begin{itemize}
    \item Defining a format incorporating all relevant information from local mapping and dynamic object detection;
    \item Investigating 2D and 3D representations;
    \item Investigating efficient neural scene graph representations;
    \item Focus on a scalable, generalizable solution.
\end{itemize}

\subsection{Implementation and Application}

After implementation, the testing and validation of the proposed perception pipeline will be carried out not only on existing datasets, but also on public roads, taking advantage of our newly built research platform, which will be introduced in this section.
In contrast to most of the existing published approaches, we strive for a full deployment on public roads rather than an adaptation to existing datasets and benchmarks.
The generated data will be published in a novel multi-modal autonomous driving dataset.

\vspace{0.5em}
\textbf{Vehicle Platform}.
Our research vehicle used for data collection and experiments is equipped with state-of-the-art hardware that is required to collect data for data-centric algorithms, such as \textit{DeepSTEP}.
The sensor setup consists of two \textit{Ouster OS-1 128}, two \textit{Innovusion Falcon}, a total of 10 cameras with different FoVs, and six RaDARs, covering a \SI{360}{\degree} field of view with multiple sensor modalities.
It will be the platform for testing and validation of the developed software and benchmarking it with the current state of the art.

\vspace{0.5em}
\textbf{Data}.
As with all deep learning-based approaches in the field of autonomous vehicle perception, the success of \textit{DeepSTEP} depends heavily on the database.
The research vehicle's setup with more sensors than required for most perception tasks allows us to generate comprehensive datasets.
On the basis of these, ablation studies will be conducted to obtain information on which sensors are important in which situations.
The dataset will be published to allow the research community to make a comprehensive comparison of different approaches.

In addition to real-world data generation, we use a simulation environment that includes a full hardware-in-the-loop (HiL-) setup to generate synthetic data.
This opens new possibilities in closing the sim-to-real gap, as already shown \cite{huch2023}.
The overall development workflow is visualized in Fig.~\ref{fig:dev_workflow}.

\section{Discussion}

The presented concept \textit{DeepSTEP} is a novel end-to-end perception pipeline that takes advantage of spatio-temporal relations in the environment.
We assume that \textit{DeepSTEP} will be able to outperform the current state of the art.
This is confirmed by recently published approaches that show the benefit of multi-model, multi-task approaches using shared information \cite{Phillips_2021_CVPR}.
These showed on the one hand the benefit of multi-modal and multi-task approaches and shared information \cite{Phillips_2021_CVPR}.
On the other hand, it has already been shown that the addition of temporal relations can improve both, computational efficiency and detection accuracy \cite{ramzy2019rst}.
Our approach will combine those approaches into a single pipeline.
This will be used as a proof of concept to further advance the approach.
Therefore, different methods for integrating temporal relations into the feature space will have to be evaluated.

An efficient sensor fusion is one of the keys for success.
We plan a feature-level fusion to obtain the maximum information content from each sensor's raw data.
This approach also allows one to quantify the contribution of each sensor to the output results.
Based on this, the sensor setup can be reduced to draw conclusions on the optimal sensor setup of autonomous vehicles.
As such an ablation study has never been conducted, the results cannot be predicted at this point, but the outcome of this study can have a huge impact on the design of future autonomous vehicles.
The success of this study is heavily dependent on the database of the experiments that will be conducted.

Data leads to the next topic of discussion.
\textit{DeepSTEP} is a heavily data-driven approach.
Thus, the performance can also be limited by a lack of data.
Due to the separated feature extraction and perception heads, we can make use of various different available datasets.
However, a focus will have to be placed on available data labels and label formats.
For example, most existing datasets do not provide labeled lane-level information.
Moreover, we plan to create our own dataset with our research vehicle and put a special focus on raw edge cases.
In general, we believe that with the presented approach, we are able to create an end-to-end perception pipeline that will generalize better to different data sources as those can be handled individually.

\section{Conclusion and Future Work}

In this work, we presented the concept of our deep learning-based spatio-temporal end-to-end perception pipeline \textit{DeepSTEP} which is currently work in progress.
A comprehensive literature review revealed the current research gap we are trying to close:
None of the existing perception algorithms fulfills all of the following requirements:
\begin{itemize}
    \item Usage of multi-modal sensor data;
    \item Integration of the temporal domain to the sensor fusion and following shared feature space;
    \item End-to-end perception with information exchange between the task-specific modules by storing information in the shared feature space.
\end{itemize}

\textit{DeepSTEP} fuses sensor-individual features in a spatio-temporal deep fusion network.
Therefore, all sensor data are abstracted to the same feature level.
This allows the addition and removal of sensor modalities to obtain information about their contributions to perception performance.
Since this shared latent space contains all information, task-specific perception heads can be applied.
Our concept comprises the usage of a detection head to detect objects and traffic participants, and a local mapping head to get road-level information about the environment.

Currently, \textit{DeepSTEP} is in the concept and proof-of-concept phase.
In the next step, we want to demonstrate the benefits of this end-to-end perception structure.
In future work, we want to bring it to public roads using our novel research vehicle.
Therefore, we plan to release a comprehensive dataset.
Encoding of all sensor data to the same level will also allow for conclusions on optimal sensor setups for autonomous vehicles.

\FloatBarrier

\flushend 
\bibliographystyle{IEEEtran}
\bibliography{main}
\end{document}